\pdfoutput=1
\documentclass[10pt,twocolumn,letterpaper]{article}
\usepackage{cvpr}
\usepackage{times}
\usepackage{epsfig}
\usepackage{graphicx}
\usepackage{amsmath}
\usepackage{amssymb}
\usepackage{soul}
\usepackage{bbm}
\usepackage{subcaption}
\usepackage{authblk}
\usepackage{multicol}
\usepackage{float}

\usepackage[pagebackref=true,breaklinks=true,letterpaper=true,colorlinks,bookmarks=false]{hyperref}

\cvprfinalcopy % *** Uncomment this line for the final submission

\begin{document}

%%%%%%%%% TITLE
\title{Fine-grained Image-to-Image Transformation towards Visual Recognition}
%\vspace{-0.1in}
\author{Wei Xiong$^{1}$ \hspace{0.1in} Yutong He$^1$ \hspace{0.1in} Yixuan Zhang$^1$ \hspace{0.1in} Wenhan Luo$^2$ \hspace{0.1in} Lin Ma$^2$ \hspace{0.1in} Jiebo Luo$^1$ \\
$^1$University of Rochester \hspace{0.2in}  $^2$Tencent AI Lab \hspace{0.2in} \\
$^1${\tt\small \{wxiong5,jluo\}@cs.rochester.edu, yhe29@u.rochester.edu, yzh215@ur.rochester.edu} \\
$^2${\tt\small \{whluo.china, forest.linma\}@gmail.com}
}
%\vspace{-0.1in}

\maketitle
\thispagestyle{empty}
%\vspace{-0.1in}

%%%%%%%%% ABSTRACT
\begin{abstract}
   Existing image-to-image transformation approaches primarily focus on synthesizing visually pleasing data. Generating images with correct identity labels is challenging yet much less explored. It is even more challenging to deal with image transformation tasks with large deformation in poses, viewpoints, or scales while preserving the identity, such as face rotation and object viewpoint morphing. In this paper, we aim at transforming an image with a fine-grained category to synthesize new images that preserve the identity of the input image, which can thereby benefit the subsequent fine-grained image recognition and few-shot learning tasks. The generated images, transformed with large geometric deformation, do not necessarily need to be of high visual quality but are required to maintain as much identity information as possible. To this end, we adopt a model based on generative adversarial networks to disentangle the identity related and unrelated factors of an image. In order to preserve the fine-grained contextual details of the input image during the deformable transformation, a constrained nonalignment connection method is proposed to construct learnable highways between intermediate convolution blocks in the generator. Moreover, an adaptive identity modulation mechanism is proposed to transfer the identity information into the output image effectively. Extensive experiments on the CompCars and Multi-PIE datasets demonstrate that our model preserves the identity of the generated images much better than the state-of-the-art image-to-image transformation models, and as a result significantly boosts the visual recognition performance in fine-grained few-shot learning. 
\end{abstract}

%%%%%%%%% BODY TEXT
%\vspace{-0.1in}
\section{Introduction}
Image-to-image transformation is an important field of visual synthesis and has many successful applications \cite{johnson2016perceptual,yu2018generative,xiong2019foreground,huang2018multimodal,zhu2016generative}. 
%It has many successful applications, such as image style transfer, image inpainting, object attribute editing and cross-modal image generation. 
A critical application of image-to-image transformation is to synthesize new images that can benefit the visual recognition systems. For example, synthesized images can augment the original training data, and subsequently boost the performance of image classification tasks \cite{antoniou2017data,wang2018low, zhang2019learning}. Synthesized images that well preserve the categorical information of the input image have been successfully applied to boost face verification \cite{yin2017multi,bao2018towards}, person re-identification \cite{ma2018disentangled} and so on. 

In this paper, we focus on fine-grained image-to-image transformation towards visual recognition, \textit{i.e.}, transforming an image with a fine-grained category to synthesize new images that preserve the identity of the input image, so that the new samples can be used to boost the performance of recognition systems. We pay special attention to transformations with large geometric deformations in object pose, viewpoint, and scale, \textit{e.g.}, face rotation \cite{hu2018pose}, caricature generation \cite{li2018carigan} and object attribute editing \cite{bao2017cvae, karras2019style} without ruining the class/identity. 
For the ultimate goal of recognition, the generated images are not necessarily required to be of high visual quality. However, they should be correctly classified even under the scenarios of fine-grained generation.
Achieving such a goal is difficult, since images from different fine-grained categories may exhibit only subtle differences. Transforming an image with geometric deformations can easily change the category of the image.

Conventional image-to-image transformation approaches primarily focus on synthesizing visually pleasing images. However, models that perform well in generating visually pleasing data do not necessarily generate identity-preserved data, thus may not benefit the subsequent recognition tasks. The problem is even more severe in fine-grained recognition because the differences between categories are inherently subtle. A possible reason is that existing generative models are not specifically designed for fine-grained image synthesis with identity preservation and visual recognition in mind.

Specifically, the performance of existing generators may be limited for the following reasons. 1) Typical generators for image-to-image transformation adopt an encoder-decoder architecture. The encoder maps the image to a condensed latent feature representation, which is then transformed into a new image by the decoder. During encoding, the latent feature fails to preserve the fine-grained contextual details of the input image, which contain rich identity information. An alternative way to preserve the contextual details is using skip-connections \cite{ronneberger2015u, he2016deep} to link feature blocks in the encoder and decoder. However, skip-connections can connect only pixels of the same spatial location in the feature blocks. It may fail on transformations with geometric deformations where there is no pixel-wise spatial correspondence between the input and output. 2) In a generator with a typical encoder-decoder architecture, the output image is decoded from the latent feature with long-range non-linear mappings. During decoding, the identity information contained in the latent feature can be weakened or even missing \cite{karras2019style}. As a consequence, the identity of the output image is not well preserved. 

To address the deformable transformation problem while maintaining contextual details, we propose a constrained nonalignment connection method to build flexible highways from the encoder feature blocks to the decoder feature blocks. With learnable attention weights, each feature point in a decoder block can non-locally match and connect to the most relevant feature points within a neighborhood sub-region of an encoder block. As such, rich contextual details from the encoder blocks can be transferred to the output image during the deformable transformation. 

To address the second problem, we propose an adaptive identity modulation method which can effectively decode the latent feature and preserve identity information. Specifically, we embed the identity feature into each convolution block of the decoder with an adaptive conditional Batch Normalization. The identity information can then be incorporated into features at different spatial resolutions and can be transferred into the output image more effectively. 

In order to generate images that better preserve the identity, we adopt a generative adversarial network (GAN) \cite{goodfellow2014generative} based framework to disentangle the identity-related factors from the unrelated factors. We apply our proposed model to two large-scale fine-grained object datasets, \textit{i.e.}, the CompCars car dataset \cite{yang2015large} and the Multi-PIE face dataset \cite{gross2010multi}. Given an image with a fine-grained category, we alter the viewpoint of the image to generate new images, which are required to preserve the identity of the input image. These generated images can benefit the few-shot learning task \cite{snell2017prototypical,finn2017model} when they are used for data augmentation.

Our primary contributions are summarized as follows.
\begin{itemize}
    \item We propose a constrained nonalignment connection method to preserve rich contextual details from the input image.
    \item We propose an adaptive identity modulation mechanism to effectively decode the identity feature to the output image so that the identity is better preserved.
    \item  Our model outperforms the state-of-the-art generative models in terms of preserving the identity and boosting the performance of fine-grained few-shot learning. 
\end{itemize}

\begin{figure*}[t]
%\vspace{-0.15in}
    \centering
    \includegraphics[width=0.98\textwidth]{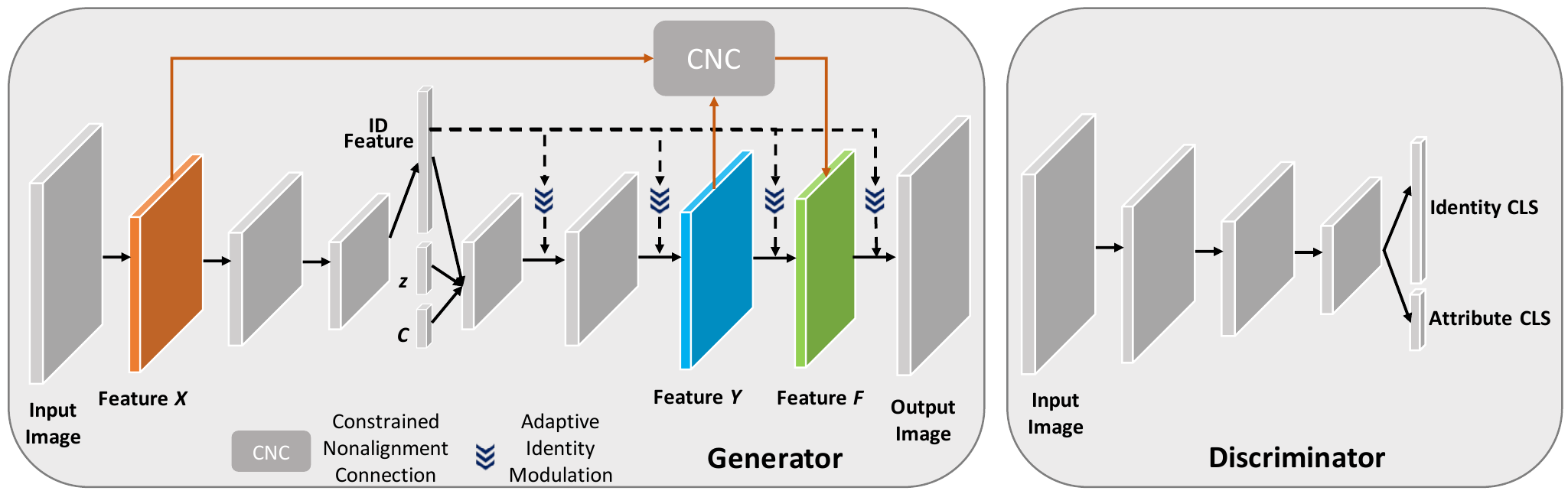}
    \caption{An overview of our model. In the generator, we use constrained nonalignment connections to preserve fine-grained contextual details from the input image, and use adaptive identity modulation to strengthen the identity information in the decoder. The discriminator outputs both the identity and attribute labels of a real or generated image (CLS: classifier).}
    \label{fig:framework}
    %\vspace{-0.1in}
\end{figure*}

%\vspace{-0.1in}
\section{Related Work}

\noindent\textbf{Generative Image-to-Image Transformation.}
Existing works have adopted conditional GANs \cite{odena2017conditional} for image-to-image transformation tasks, such as image inpainting \cite{yu2018generative, xiong2019foreground}, super-resolution \cite{ledig2017photo}, and general-purpose image-to-image translation tasks \cite{isola2017image, zhu2017unpaired}. Many models mainly handle scenarios where the input image and output image have pixel-wise spatial correspondence, and tend to fail on geometric transformation tasks, which are specifically addressed by our work. Recent works have made attempts on geometric transformation tasks, including object rotation and deformation learning with spatial transformer networks \cite{jaderberg2015spatial} and deformable convolution  \cite{dai2017deformable}, face viewpoint rotation \cite{tran2017disentangled, huang2017beyond}, person generation with different poses \cite{ma2018disentangled, ma2017pose} and vehicle generation with different viewpoints \cite{zhu2018visual,nguyen2019hologan}. 

However, existing works primarily aim at synthesizing data of high visual quality \cite{karras2017progressive,karras2019style,brock2018large,zhang2018self,wang2018high}. They are not specifically designed to preserve the identity of the generated images, especially under the scenarios of fine-grained image transformation, which is our primary goal. For example, StyleGAN \cite{karras2019style} and PG-GAN \cite{karras2017progressive} can generate high-quality faces, but the faces have no identity labels. Several works can synthesize fine-grained categorical images \cite{bao2017cvae}. However, they are directly conditioned on category labels, which thereby cannot generalize to new categories.

Our work differs from the conventional image transformation works in the following aspects. 1) Our primary goal is to synthesize images with a correct identity so that the generated images can benefit the subsequent fine-grained recognition tasks. Our model is specifically designed for preserving the fine-grained details that can benefit identity preservation. We emphasize that high visual quality is {\it not} necessarily required for identity preservation. 2) We address the task of image-to-image transformation with large geometric deformations. There is no pixel-wise correspondence between the input and the output images. 3) Our model can generalize to unseen categories. Therefore it can benefit the few-shot learning task by augmenting the data in new categories. 

 \noindent\textbf{Non-Local Networks.}
 Our proposed constrained nonalignment connection is related to non-local networks. The idea of non-local optimization has been proposed and used in many traditional vision tasks, such as filtering and denoising \cite{buades2005non,dabov2007image}. Recently, such an idea has been extended to neural networks to compute the long-range dependencies within feature maps, such as non-local neural networks \cite{wang2018non,liu2018non, chen20182} and self-attention GAN \cite{zhang2018self}. 
 Our model differs from the existing non-local structure in two aspects. First, we compute non-local attention between different feature maps to construct information highways in a deep generator, while existing models typically calculate attention within the same feature, \textit{i.e.}, self-attention. Second, conventional non-local structures usually calculate the attention in the whole searching space, which may be challenging to optimize. On the contrary, our proposed constrained nonalignment connection reduces the non-local searching scope to capture the feature correspondences more effectively. 
 
 \noindent\textbf{Network Modulation.}
 Network modulation is a technique that modulates the behavior of network layers with a given conditioning feature \cite{de2017modulating}.  It has been proved effective in several tasks \cite{yang2018efficient, wang2018recovering, perez2018film, strub2018visual,chen2018self,karras2019style}. It is typically realized by mapping the conditioning feature to the hidden variables of a layer, such as the re-scale factors of Batch Normalization \cite{de2017modulating} or Instance Normalization \cite{karras2019style}. In our work, a novel modulation method is proposed to regularize the convolution layers by adaptively integrating the identity feature and the convolutional feature maps.

\vspace{-0.1in}
\section{Our Approach}
As shown in Fig. \ref{fig:framework}, our model is composed of a generator $G$ and a discriminator $D$. The generator takes an image $I$, random noise $z$ and a condition code $C$ as inputs, and generates a new image $I_f$. $C$ is a vector encoding an attribute of an image, such as viewpoint or pose. The discriminator predicts both the identity and attribute class probabilities of an image. The identity of $I_f$ is required to be the same as that of input image $I$, \textit{i.e.}, identity preservation.

\subsection{Generator}
Our generator adopts an encoder-decoder architecture, \textit{i.e.}, $G=\{Enc, Dec\}$. The encoder $Enc$ maps the input image to an identity feature vector $f_{id} = Enc(I)$, which is then concatenated with noise $z$ and target attribute code $C$ to form the latent vector $f_l=cat[f_{id}, z, C]$. The latent vector is then decoded by the decoder $Dec$ to the output image $I_f = Dec(f_l)$. To preserve the contextual details of the input image during deformable transformation, we propose a constrained nonalignment connection $CNC(X,Y)$ that can link the intermediate feature map $X$ in the encoder and feature map $Y$ in the decoder with non-local attention maps. To better preserve the identity, we propose an adaptive identity modulation method to effectively embed the identity feature $f_{id}$ into the convolution blocks of the decoder.

\begin{figure}[t]
%\vspace{-0.1in}
    \centering
    \includegraphics[width=0.44\textwidth]{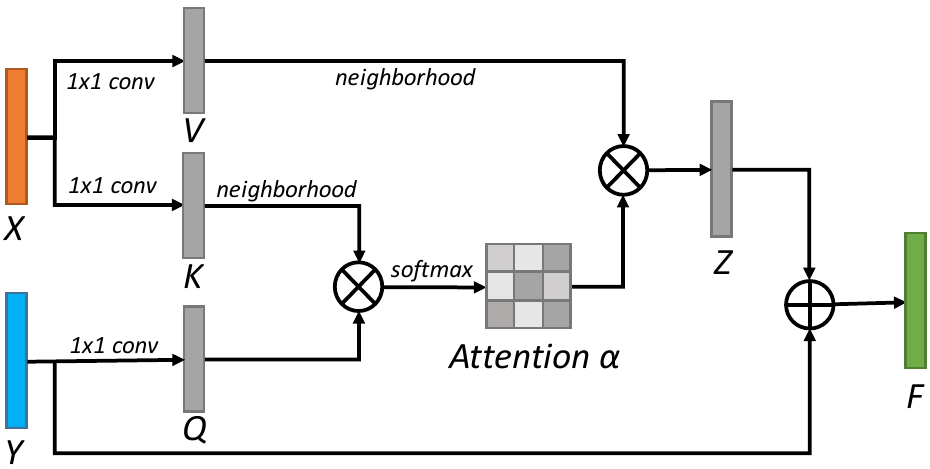}
    \caption{Structure of constrained nonalignment connection. $\otimes$ denotes matrix multiplication. $\oplus$ denotes concatenation.}
    \label{fig:attention}
    %\vspace{-0.1in}
\end{figure}

%\vspace{-0.1in}
\subsubsection{Constrained Nonalignment Connection}

Fig. \ref{fig:attention} shows the structure of our constrained nonalignment connection. Consider an intermediate feature map $X\in \mathbb{R}^{C_X\times H_X \times W_X}$ in the encoder and an intermediate feature map $Y\in \mathbb{R}^{C_Y\times H_Y \times W_Y}$ in the decoder. (We ignore the batch size for simplicity.) Feature $Y$ may lose fine-grained contextual details that are complementary for identity preservation during layers of mapping in the generator \cite{he2016deep}. To address this issue, we selectively link $Y$ and $X$ with a non-local attention map, so that the attended feature $Z$ contains rich contextual details from $X$. At the same time, the generator still learns a correct geometric transformation. 

Specifically, we first reshape the feature $X$ to the shape $C_X\times{N_X}$, where $N_X=H_X\times{W_X}$. Similarly, we obtain the reshaped feature $Y\in \mathbb{R}^{C_Y\times{N_Y}}$.
We then use several $1\times1$ convolutions to project $X$ into key $K\in \mathbb{R}^{C_h\times C_X}$, value $V\in \mathbb{R}^{C_h\times C_X}$ and $Y$ into query $Q\in \mathbb{R}^{C_h\times C_Y}$, so that they are in the same feature space.

\begin{figure}[t]
%\vspace{-0.1in}
    \centering
    \includegraphics[width=0.4\textwidth]{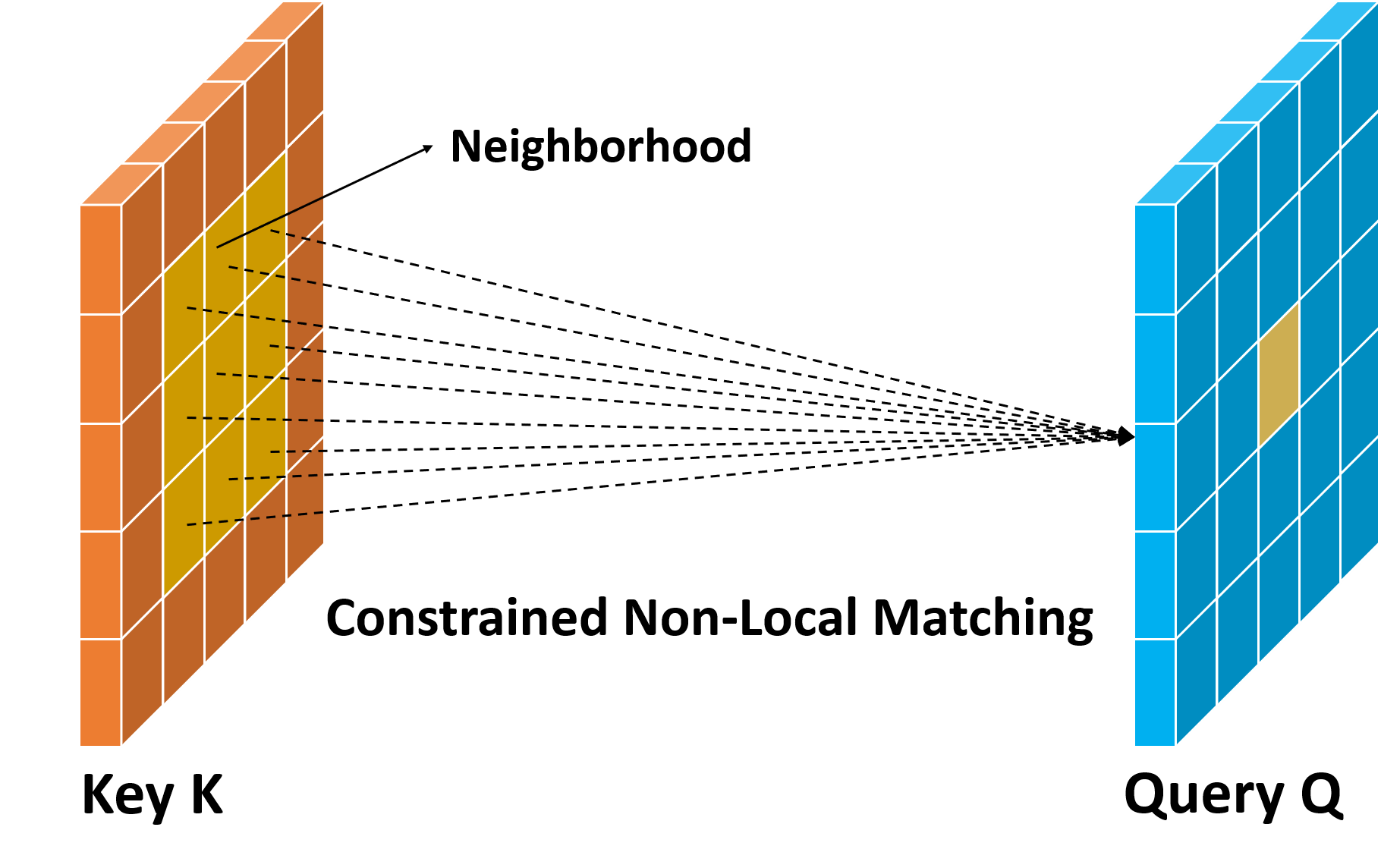}
    \caption{Illustration of constrained non-local matching between query $Q$  and key $K$. Each feature point in $Q$ can non-locally match over all the spatial locations inside a neighborhood region of $K$. }
    \label{fig:nonalignment_toy}
    %\vspace{-0.1in}
\end{figure}

Next, for each spatial location $p$ in $Q$, we use the feature point $Q_p$ to attend to the feature points in $K$ and obtain a non-local attention map $\alpha_p$. Conventional non-local networks typically calculate the attention map by matching $Q_p$ with features of all the spatial locations in $K$, which is both time-consuming and difficult to optimize. Considering a point in the input image, in most situations, after the geometric transformation, the spatial location of that point is usually changed within a certain neighborhood region around the point. Inspired by this observation, we propose a \textit{constrained non-local matching} between the query $Q$ and the key $K$. As shown in Fig. \ref{fig:nonalignment_toy}, for each spatial location $p$ in $Q$, we define a corresponding neighborhood region in $K$ as $\mathcal{N}_p$, which is a square area with its center at location $p$. We define the radius of the neighborhood with a hyper-parameter $r$, then the spatial size of the neighborhood region is $(2r+1)\times(2r+1)$. For each location $p$, we extract all the features in neighborhood $\mathcal{N}_p$ from feature $K$, denoted as $K_{\mathcal{N}_p}\in \mathbb{R}^{C_h\times (2r+1)(2r+1)}$, then use $Q_p$ to attend to $K_{\mathcal{N}_p}$ and calculate the constrained non-local attention as
\begin{equation}
    \alpha_p=Q_p^TK_{\mathcal{N}_p}.
    \label{eq:attention}
\end{equation}
We normalize $\alpha_p$ using the softmax function so that the weights are summed to 1. Feature at location $p$ of the attended feature $Z$ is the weighted sum over all the feature points in neighborhood $\mathcal{N}_p$ of the value $V$, formulated as $Z_p=\sum_{i\in \mathcal{N}_p}\alpha_{p}^i V_{\mathcal{N}_p}^i$.
 We then concatenate the attended feature with the original feature $Y$, to obtain the final fused feature $F = [Y,Z]$.

%\vspace{-0.1in}
\subsubsection{Adaptive Identity Modulation}

In the decoder, directly mapping the latent feature to an image with layers of convolution may not be optimal. During the long-range mapping, the identity information may be weakened \cite{he2016deep, karras2019style} or missing. To address this problem, we propose an adaptive identity modulation method to transfer identity information to the output image effectively.

Specifically, we embed the identity feature into the convolution blocks, so that feature maps at each spatial resolution can perceive and utilize the identity knowledge. To this end, we adopt conditional batch normalization \cite{de2017modulating} to modulate the convolution layers \cite{de2017modulating,perez2017learning} with identity feature $f_{id}$. 
Given a mini-batch of features $\{B_{i,:,:,:}\}_{i=1}^{N}$ in a convolution layer, we first normalize each feature $B_i$ with Batch Normalization (BN) \cite{ioffe2015batch}

\begin{equation}
    \hat{B}_{i} = \frac{B_{i}-\mathbb{E}[B_{:,c,:,:}]}{\sqrt{var[B_{:,c,:,:}]+\epsilon}},
\end{equation}
where $B_{i}$ is the feature map of the $i$-th sample in the batch,  $\epsilon$ is a constant for numerical stability. In the vanilla BN,  we re-scale the feature with two learnable parameters $\gamma$ and $\beta$. 

In order to better decode the identity feature, we adopt a conditional Batch Normalization (CBN) to learn the re-scale parameters $\gamma$ and $\beta$ on condition of the identity feature $f_{id}$. Then in each convolution block, we have
\begin{equation}
    \tilde{B}_{i} = \gamma(f_{id}) \hat{B}_{i} + \beta(f_{id}) ,
    \label{eq:im}
\end{equation}
where $\gamma(f_{id})$ and $\beta(f_{id})$ are functions of $f_{id}$. 

In traditional CBN, the re-scale parameters $\gamma$ and $\beta$ usually depend only on the conditioning feature. However, we argue that different feature maps should perceive the conditioning feature in different ways. Features in different convolution layers exhibit different functionalities and may pay different attention to the conditioning feature. In order to adaptively perceive and integrate the conditioning feature, we re-formulate $\gamma$ and $\beta$ to be conditioned on both the feature map to be modulated and the conditioning feature:

\begin{equation}
    \tilde{B}_{i} = \gamma(f_{id}, B_i) \hat{B}_{i} + \beta(f_{id}, B_i) ,
    \label{eq:aim}
\end{equation}
where $\gamma(f_{id}, B_i)$ and $\beta(f_{id}, B_i)$ are functions of $f_{id}$ and $B_i$.

Specifically, we first calculate the average feature $B_f$ of $B_i$ over spatial locations, \textit{i.e.}, $B_f = \frac{1}{H\times W}\sum_{h,w} B_{i,:,h,w}$. Then we calculate an attention using $B_f$, formulated as $att_B=\tau(B_f)$, where $\tau$ can be realized with a MLP composed of several dense layers with the activation of the last layer to be Sigmoid.
We obtain the attended feature as: 
\begin{equation}
    f_{id}^{att} = f_{id} \odot att_B,
\end{equation}
where $\odot$ denotes element-wise multiplication. As such, the identity feature is adaptively selected by the feature map $B_i$. 

The attended identity feature $f_{id}^{att}$ is then mapped to $\gamma$ and $\beta$ with two MLPs. By embedding the identity features into convolution layers on condition of the features to be modulated, the identity-related information can be better integrated by the decoder. The detailed structure of our adaptive identity modulation is in the supplementary materials. 

%\vspace{-0.1in}
\subsection{Discriminator and Objective Functions}
To encourage the model to generate identity-preserved images, our discriminator $D$ adopts a similar architecture as ACGAN \cite{odena2017conditional}. $D$ is composed of several convolution blocks, followed by an identity classification layer $D^i$, and an attribute classification layer $D^a$.

We denote $y_a^t$ as the target attribute label, which can be encoded into the one-hot code $C$. During training, the identity label $y_i$ and the attribute label $y_a$ of the input image $I$ are provided to train the classifier in $D$, where $1\leq y_i\leq N_i$ and $1\leq y_a\leq N_a$. $N_i$ and $N_a$ are the number of identity and attribute categories in the training data, respectively. 

Upon training the discriminator, we assign the ground-truth identity label of the fake image $I_f$ as $N_i+1$. In this way, the discriminator can not only classify the real image but also distinguish the real image from the fake one. We use the following objective to optimize $D$:

%\vspace{-0.15in}
\begin{equation}
%\vspace{-0.1in}
    \begin{split}
    \underset{D}{\max}\ J(G,D) &= \mathbb{E}[\log D^i_{y_i}(I)]+\mathbb{E}[\log D^i_{N_i+1}(G(I))] \\
    &+ \lambda\mathbb{E}[\log D^a_{y_a}(I)],
    \end{split}
    %\vspace{-0.1in}
\end{equation}
where $J$ is the value function, $D^i_k$ and $D^a_k$ are the $k$-th element in $D^i$ and $D^a$, respectively. $\lambda$ is a hyper-parameter to balance the weight of identity classification and attribute classification.

When training the generator, we encourage the generated image to have the same identity label $y_i$ as the input image as well as the target attribute label $y_a^t$ by optimizing the following objective:
%\vspace{-0.1in}
\begin{equation}
    \underset{G}{\max}\ J(G,D) = \mathbb{E}[\log D^i_{y_i}(G(I))] + \lambda\mathbb{E}[ \log D^a_{y_a^t}(G(I)) ].
\end{equation}

%Optimized with the two classification objectives, our model is able to generate identity-preserved images that also correctly reflect the input attribute. 

%\vspace{-0.1in}
\section{Experiments}
We evaluate our model on two challenging datasets, CompCars dataset \cite{yang2015large} and Multi-PIE dataset \cite{gross2010multi}. CompCars dataset contains over 1,700 categories of car models and 100,000 images. Multi-PIE dataset contains face images of  337 identities. Both datasets  are {\it quite large for fine-grained image generation and few shot learning}. 
We perform viewpoint morphing on both datasets. Given an image, a target viewpoint, and random noise, our goal is to generate new images belonging to the same identity/model category as the input image with the target viewpoint. 
We conduct two types of experiments. The first one is identity preservation. In this experiment, we derive a classifier on the real images, which are then used to classify the generated images. 
%We choose KNN classifier as it is parameter-free, so that it can straightforwardly reveal the classification performance. 
The second type is few-shot learning. In this experiment, we use the generated images to augment the training data and test how the generative models can benefit the performance of the few-shot classifier.

\subsection{Experiment Settings}
\textbf{Dataset.} For Multi-PIE dataset, following the setting in \cite{tran2017disentangled}, we use $337$ subjects with neutral expression and $9$ poses within $\pm60$ degree. The first $200$ subjects form an auxiliary set, which is used for training the generative models. The rest $137$ subjects form a standard set, which is used to conduct visual recognition experiments. We crop and align the faces and resize each image to $96\times96$.

The car images in the CompCars \cite{yang2015large} dataset contain several viewpoints, including frontal, frontal left side, rear view, rear side, side, and other views. Note that the same car model can have totally different colors. Since the rear views may contribute less to the identification of the car model, we remove all the images with rear views and keep only images with the following five viewpoints: frontal, frontal left, frontal right, left side, and right side. We also remove minor categories containing less than $10$ samples. All the images are resized to $224\times224$. Similar to the setting in Multi-PIE, we split the filtered dataset into an auxiliary set which contains images of $1,181$ car models, and a standard set which contains images of another $296$ car models. These two sets are disjoint in terms of model category.

\noindent\textbf{Existing Models to Compare.}
We compare our model with the state-of-the-art models DR-GAN \cite{tran2017disentangled}, CR-GAN \cite{tian2018cr} and Two-step \cite{hadad2018two}, which also aim at generating fine-grained objects given a target attribute as the condition. For a fair comparison, we adjust the generator of each model to have a comparable amount of parameters. Note that there are other models for image-to-image transformation. However, many of them need pose masks or landmarks as guidance \cite{li2018carigan,ma2018disentangled}, which differs from our setting. Therefore, it is not appropriate to compare them with our model. We also do not compare our model with StyleGAN \cite{karras2019style}, PG-GAN \cite{karras2017progressive}, or other similar models since they are unconditional models that cannot generate categorical images. 

\noindent\textbf{Evaluation Metric.}
Since our task is visual recognition oriented image transformation, we primarily evaluate the identity preservation performance of each model and report classification accuracy on identity preservation and few-shot learning experiments. We do not use FID \cite{heusel2017gans} or Inception Score \cite{salimans2016improved} to quantify the generated images since they are mainly used to evaluate the visual quality of images. 

\begin{figure}[t]
%\vspace{-0.1in}
\centering
\begin{tabular}{c}
{\includegraphics[width=0.4\textwidth]{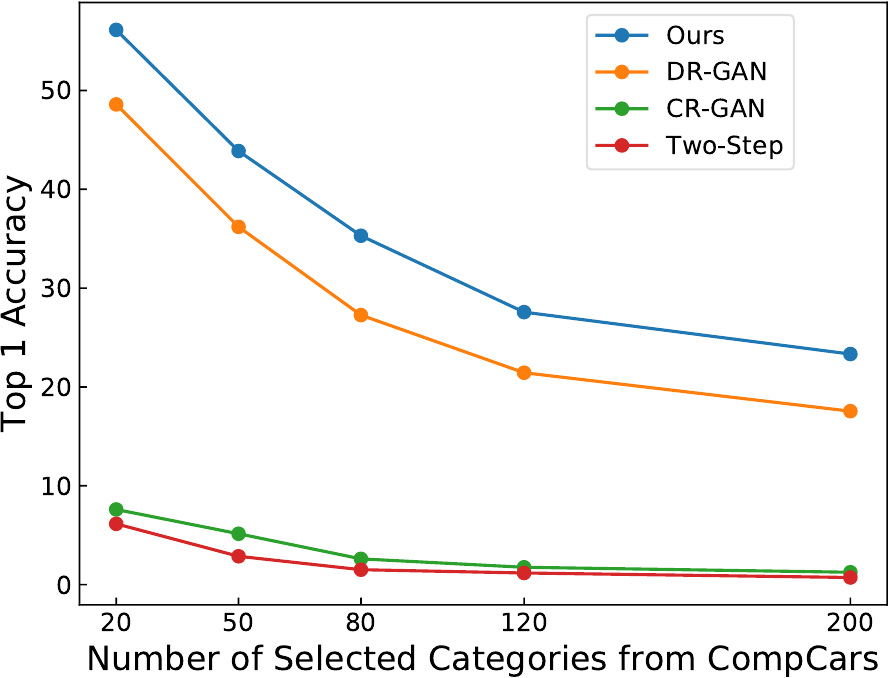}} \\
{\includegraphics[width=0.4\textwidth]{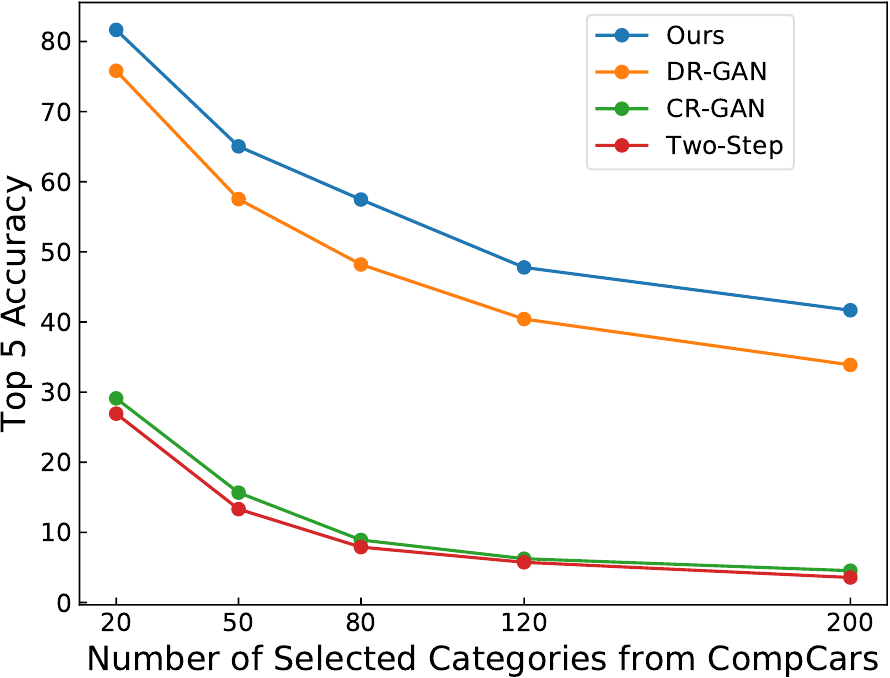}} \\
%(a) Top 1 Accuracy.  & (b) Top 5 Accuracy.
\end{tabular}
\caption{Classification accuracy on the generated images from CompCars dataset with $20$, $50$, $80$, $120$ and $200$ categories. }
\label{fig:knn_acc}
%\vspace{-0.15in}
\end{figure}

\noindent\textbf{Implementation Details. }
Our model is optimized with Adam optimizer. The learning rate is $0.0002$, and the batch size is $64$. On CompCars dataset, in each training cycle, we train one step for the generator and one step for the discriminator. The target viewpoint code $C$ is a $5\times1$ one-hot vector. We empirically choose the radius of neighborhood $r=7$ for feature maps with size $28\times28$ and $r=14$ for feature maps with size $56\times56$. We set $\lambda$ to be $5$. On Multi-PIE dataset, we train four steps for the generator and one step for the discriminator in each training cycle. The target viewpoint code is a $9\times1$ one-hot vector. We empirically choose the radius of neighborhood $r=6$ for feature maps with size $24\times24$. The noise vector has a size of $128\times1$.We set $\lambda$ as $1$.

%\vspace{-0.05in}
\subsection{Identity Preservation}
In this section, we evaluate the identity preservation ability of each generative model on both CompCars and Multi-PIE datasets. On each dataset, we first train each model on the whole auxiliary set to learn the viewpoint transformation. We also train a Resnet18 \cite{he2016deep} model on the auxiliary set, then use its features of the last pooling layer as the representation for identity classification experiments.

On CompCars dataset, we select $N_c$ car models from all the $296$ classes in the standard set and choose all the images in the selected $N_c$ classes to form the dataset on which the classification experiment will be conducted. We randomly split the selected dataset as train and test sets with a ratio of $8:2$. Note that the train and test sets contain images from all the $N_c$ classes. We train a KNN classifier on the train set with the Resnet18 model as the feature extractor. Following that, for each image in the test set, we transform it with the generative model, which outputs five images, one per specific target viewpoint. We then use the KNN classifier to classify all the generated images and report the top-1 and top-5 accuracies of each model. We choose the KNN classifier because it is parameter-free so that it can directly reveal the separability of the generated samples.

\begin{figure}[t]
%\vspace{-0.1in}
\centering
\begin{tabular}{c}
{\includegraphics[width=0.4\textwidth]{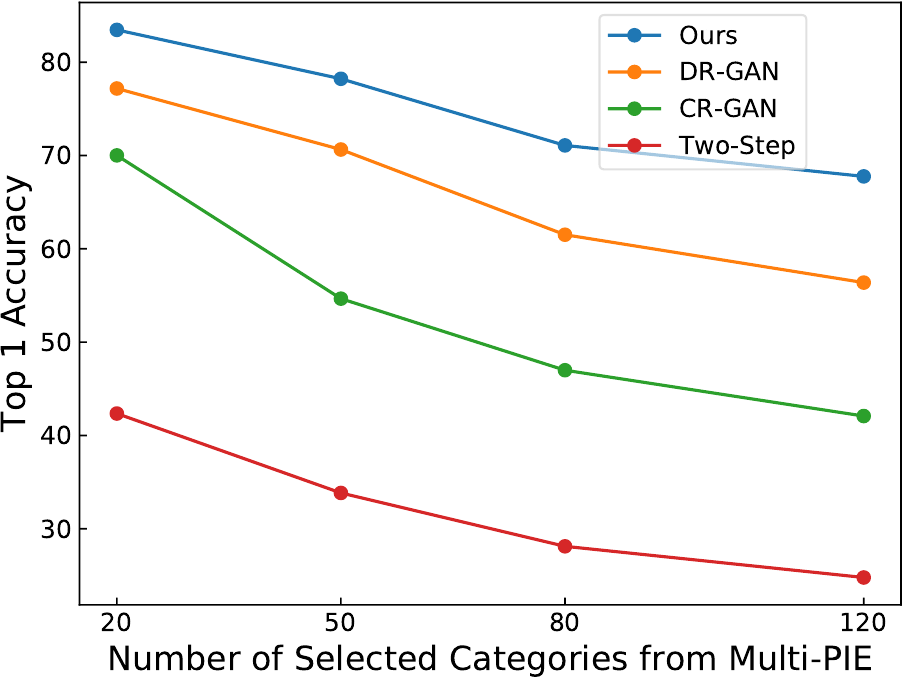}} \\
{\includegraphics[width=0.4\textwidth]{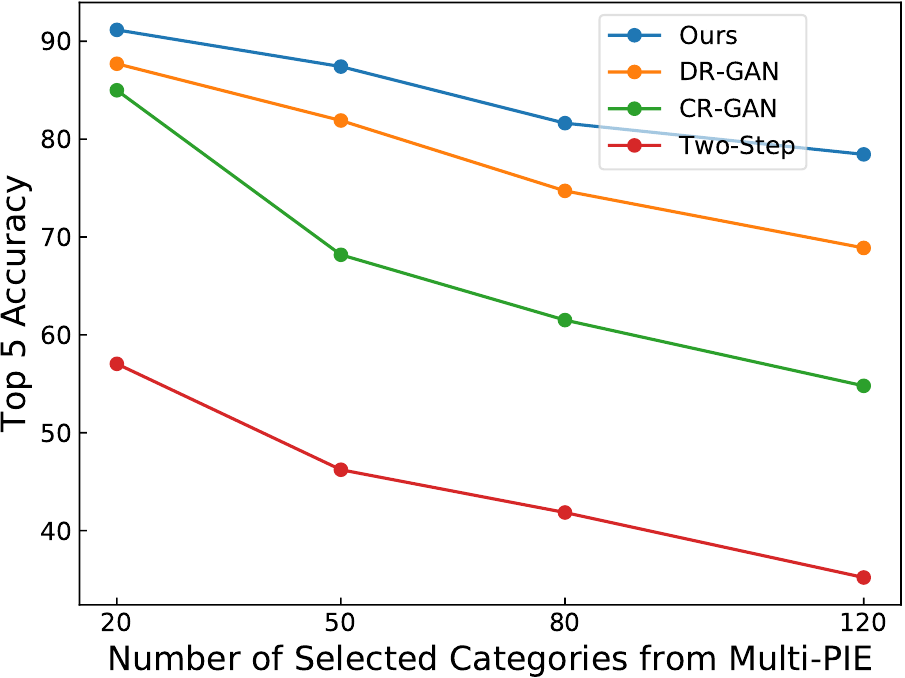}} \\
%(a) Top 1 Accuracy.  & (b) Top 5 Accuracy.
\end{tabular}
\caption{Classification accuracy on the generated images from Multi-PIE dataset with $20$, $50$, $80$ and $120$ categories.}
\label{fig:knn_acc2}
%\vspace{-0.1in}
\end{figure}

Fig. \ref{fig:knn_acc} shows the results of each model. We select $N_c$ to be $20$, $50$, $80$, $120$ and $200$. Our full model with both CNC and AIM  significantly outperforms the existing models by a large margin (over \textbf{5\%} accuracy gain under all settings), showing that our model can better preserve the identity of the generated images.

We conduct a similar identity preservation experiment on Multi-PIE dataset, except that we select $N_c$ to be $20$, $50$, $80$ and $120$ from $137$ classes in the standard set and generate $9$ fake images (viewpoints ranging from $-60$ degree to $60$ degree) from each input test image. Fig. \ref{fig:knn_acc2} shows the classification results of each model on the generated face data. Our model again outperforms the existing models, further demonstrating the superiority of our model. 

\begin{figure}[t]
%\vspace{-0.15in}
    \centering
    \includegraphics[width=0.46\textwidth]{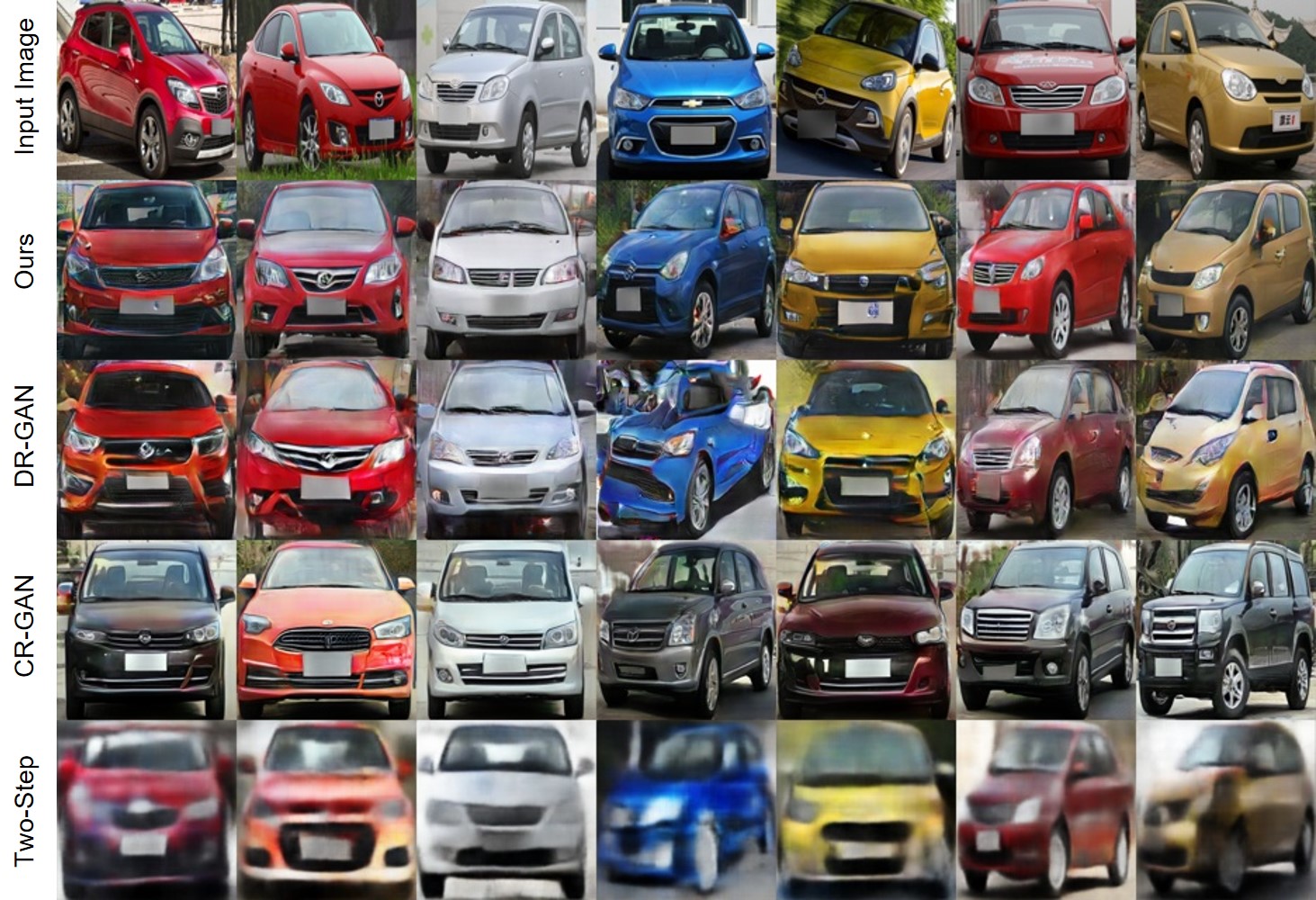}
    \caption{Exemplar images generated by different models on CompCars dataset. In each column, from the top to the bottom are: input image, and results of our model, DR-GAN \cite{tran2017disentangled}, CR-GAN \cite{tian2018cr}, Two-Step \cite{hadad2018two}, respectively. 
    Since all the models generate the correct viewpoints, we do not show the viewpoints here. 
    }
    \label{fig:quality}
    %\vspace{-0.1in}
\end{figure}

\begin{figure}[t]
%\vspace{-0.15in}
    \centering
    \includegraphics[width=0.48\textwidth]{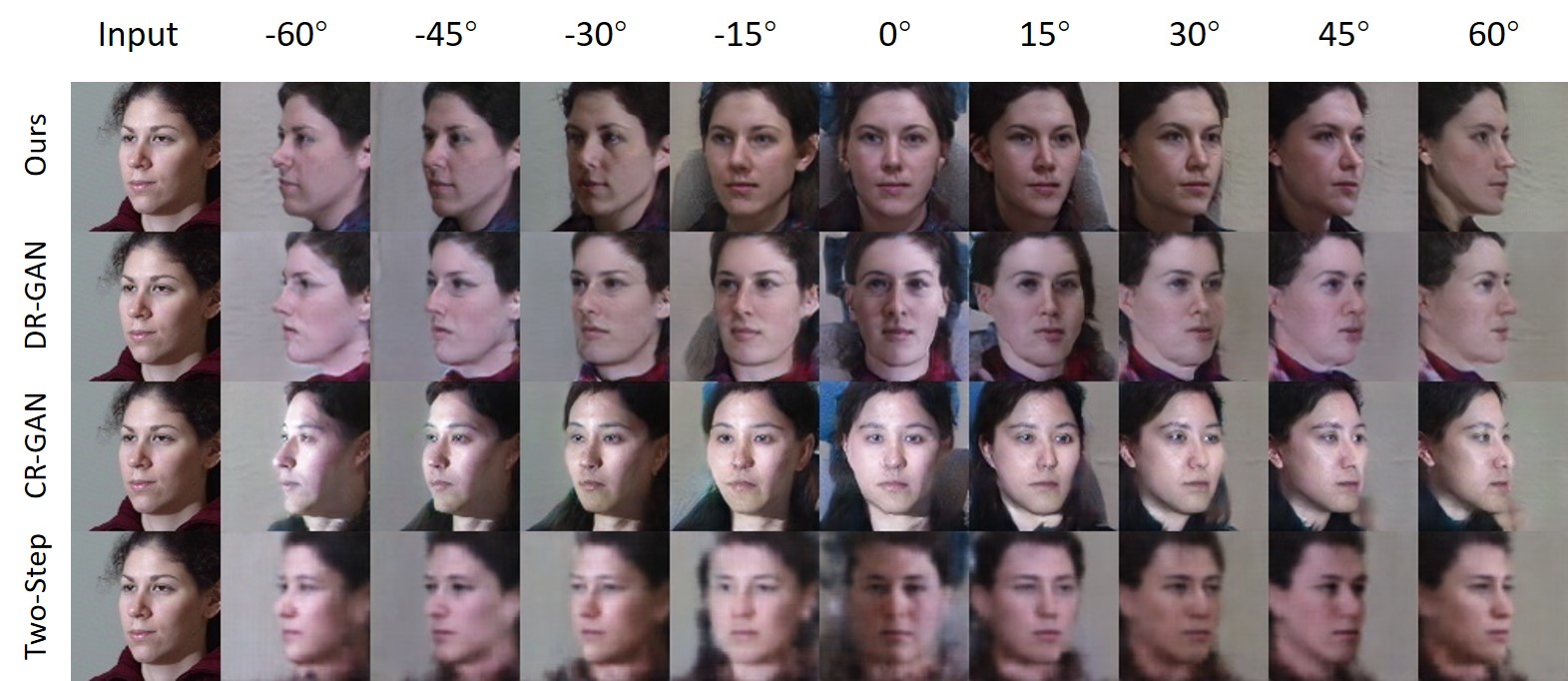}
    \caption{Exemplar images generated by different models on Multi-PIE dataset. From left to the right: input image, generated images with 9 different viewpoints. Please pay attention to the details such as face shape, hair and mouth.
    }
    \label{fig:quality_multipie}
    %\vspace{-0.05in}
\end{figure}

To make a more thorough analysis of the results, we investigate each model by showing their visual results straightforwardly, as shown in 
Fig. \ref{fig:quality} on CompCars dataset and Fig. \ref{fig:quality_multipie} on Multi-PIE dataset. 

Seen from Fig. \ref{fig:quality}, DR-GAN, CR-GAN and our model can generate sharp images, while Two-Step can only generate blurry images. Although images generated by CR-GAN look realistic, the key regions that identify a car (such as bumper and lights) are quite different from the input image, showing that their identity is not well preserved. This observation is consistent with the classification performance in Fig. \ref{fig:knn_acc}. The results further indicate that high-quality images do not necessarily stand for identity-preserved images. Our model can generate fine-grained details that are almost in accordance with the input image. Note that in some situations, our model fails to capture all the details of the input car. It is because we are dealing with fine-grained image transformation with large deformation, which is very challenging. Moreover, cars in our dataset contain many details, making the task more difficult to accomplish. Even though, images generated by our model still preserve many more details than all the existing methods, demonstrating the effectiveness of our model.  

Fig. \ref{fig:quality_multipie} shows an exemplar case from Multi-PIE dataset. We input the same image to the generative models, outputting images with nine different viewpoints. DR-GAN, CR-GAN and Two-Step fail to preserve the identity very well. On the contrary, our model can generate images whose identity is almost the same as the input image, with as many details preserved as possible, demonstrating the effectiveness of our model in identity preservation.

%We further compare each model by visualizing the t-Distributed Stochastic Neighbor Embedding (t-SNE) \cite{hinton2017visualizing} plot on their generated images. We only compare our model with DR-GAN, since it has the best identity preservation ability among all the models we have compared. We generate images of 10 classes with our model/DR-GAN and use the Resnet18 model to extract features of each image, then plot the images. As can be seen from Fig. \ref{fig:tsne}, our model can generate categorical data with smaller intra-class variance and larger inter-class variance compared to DR-GAN, further demonstrating the superiority of our model on identity preservation.

%\vspace{-0.05in}
\subsection{Few-shot Learning}
In this section, we evaluate how well each generative model can boost the performance of the fine-grained few-shot learning task \cite{li2019revisiting} when used as a data augmentation method. Experiments are conducted on the CompCars dataset. Similar to the identity preservation experiment, we train the generative models on the whole auxiliary set. 
%In this experiment, we do not fine-tune the Resnet18 model on the auxiliary set. 

We randomly select $N_c$ car models from all the $296$ model classes in the standard set. Then we select images of the $N_c$ classes to form a selected dataset on which we will conduct the few-shot learning experiment. We randomly select $s$ images from each car model ($N_c$ car models in total) to form the few-shot train set, and use all the rest images as the test set. Under such a setting, the few-shot classification task can be named as ``$N_c$ way $s$ shot'' few-shot learning. 

In this experiment, we adopt Resnet18 as the classifier \footnote{the last layer of Resnet18 is modified to $N_c$ nodes.} for few-shot learning. We first train the classifier only on the train set, which is then used to classify the images in the test set. Different from the setting in the identity preservation experiment,  we classify the real images instead of the fake images. We then input the images in the train set to the generative model and generate 20 fake images per image in the few-shot train set and set their identity labels to be the same as the input image. To generate diverse images, we interpolate between different viewpoint codes and input the new code to the generator as the target viewpoint. The generated images are used to augment the train set. 

We then retrain the Resnet18 on the augmented train set and classify images in the original test set. Note that when training the Resnet18 classifier with the augmented data, we also input the real/fake label to the Resnet18, so that the model can balance the importance of generated data and real data. Specifically, when training the Resnet18 with a real image, we also input the label $1$ (a 1-bit vector concatenated with the feature of global pooling layer in Resnet18) to the model. When training the Resnet18 with a fake image, we input label $0$ to the model. During testing, since the test images are all real images, we input the label $1$ along with the image to the classifier, to obtain the prediction.

\begin{table}[t]
%\vspace{-0.05in}
    \centering
    %\small
    \setlength\tabcolsep{0.3cm}
    
    \begin{tabular}{@{}lcc@{}}%{|C{2.5cm}|C{1.8cm}|C{1.8cm}|C{1.8cm}|}
        \hline
        Model & 20w-5s \% & 20w-10s \%  \\
        \hline
        $w$/$o$  augment &29.77 &55.86   \\
        augment + Two-Step  &32.04 &52.53  \\
        augment + CR-GAN  &27.61 &39.67  \\
        augment + DR-GAN  &47.85 &60.01  \\
        augment + Ours  &\textbf{52.44} &\textbf{66.93}  \\
        \hline
    \end{tabular}
     \caption{ Classification accuracy of few-shot learning under different settings on CompCars dataset. ``$m$w-$n$s" means $m$ way $n$ shot learning. ``$w$/$o$'' denotes ``without''. }
    \label{table:fewshot}
    %\vspace{-0.1in}
\end{table}

\begin{table*}[t]
%\vspace{-0.1in}
    \centering
    %\small
    %\setlength\tabcolsep{0.cm}

    \begin{tabular}{lcccccc}%{|C{2.5cm}|C{1.8cm}|C{1.8cm}|C{1.8cm}|}
        \hline
        Model & 20c-top1 \% & 20c-top5 \% &  50c-top1 \% & 50c-top5 \% &80c-top1 \% & 80c-top5 \% \\
        \hline
        vanilla &48.59 &75.82 &36.20 &57.52 &27.27 &48.20 \\
        vanilla + Deformable Conv \cite{dai2017deformable}  &49.75 &76.08 &37.26 &58.53 &28.82 &48.81 \\
        vanilla + Global-NC(56) &50.37 &76.25 & 37.45 &58.21 &29.23& 49.39\\
        vanilla + CNC(56) &52.45 &78.31 & 39.42 &60.52 &31.38& 52.88\\
        vanilla + Global-NC(28) &53.12 &77.08 & 38.30 &59.12 &30.40& 52.13 \\
        vanilla + CNC(28) &55.05 &80.16 &42.24 &63.49 &34.68 &56.09 \\
        vanilla + CNC(28) + IM (Eq. (\ref{eq:im})) &55.47 &81.22 &42.35 &64.73 &34.92 &56.80 \\
        vanilla + CNC(28) + AIM (Eq. (\ref{eq:aim}))   &\textbf{56.13} &\textbf{81.65} &\textbf{43.87} &\textbf{65.04} &\textbf{35.30} &\textbf{57.46} \\
        \hline
    \end{tabular}
    \caption{ Identity preservation experiment results with different versions of our model on CompCars dataset. Experiments are done with  20, 50, and 80 categories from the standard set. We report both top-1 and top-5 accuracies. 
    %``CNC(28)'' refers to the constrained nonalignment connection applied to $28\times28$ feature blocks, ``Non-CNC'' refers to unconstrained nonalignment connection and ``IM'' refers to identity modulation. 
    }
    \label{table:ablation}
    %\vspace{-0.1in}
\end{table*}

We report the few-shot learning results boosted by different generative models under $N_c$ classes, where $N_c=20$ in our experiment. As shown in Table~\ref{table:fewshot}, without any augmented data, training on limited real samples leads to poor performance on the test data. Using the generated images by our model or DR-GAN to augment data can significantly boost the performance of the classifier, indicating that it is an effective way to boost the few-shot learning by augmenting the data with generative models. Our model yields much better performance than DR-GAN. Interestingly, since the images generated by CR-GAN and Two-Step do not well preserve the identity, using them to augment data does not benefit the few-shot classification. The results indicate that generators with better identity preservation ability  lead to more significant improvements in few-shot learning, while weak generators can even hurt the performance.

\begin{figure}[t]
    %\vspace{-0.1in}
    \centering
    \includegraphics[width=0.48\textwidth]{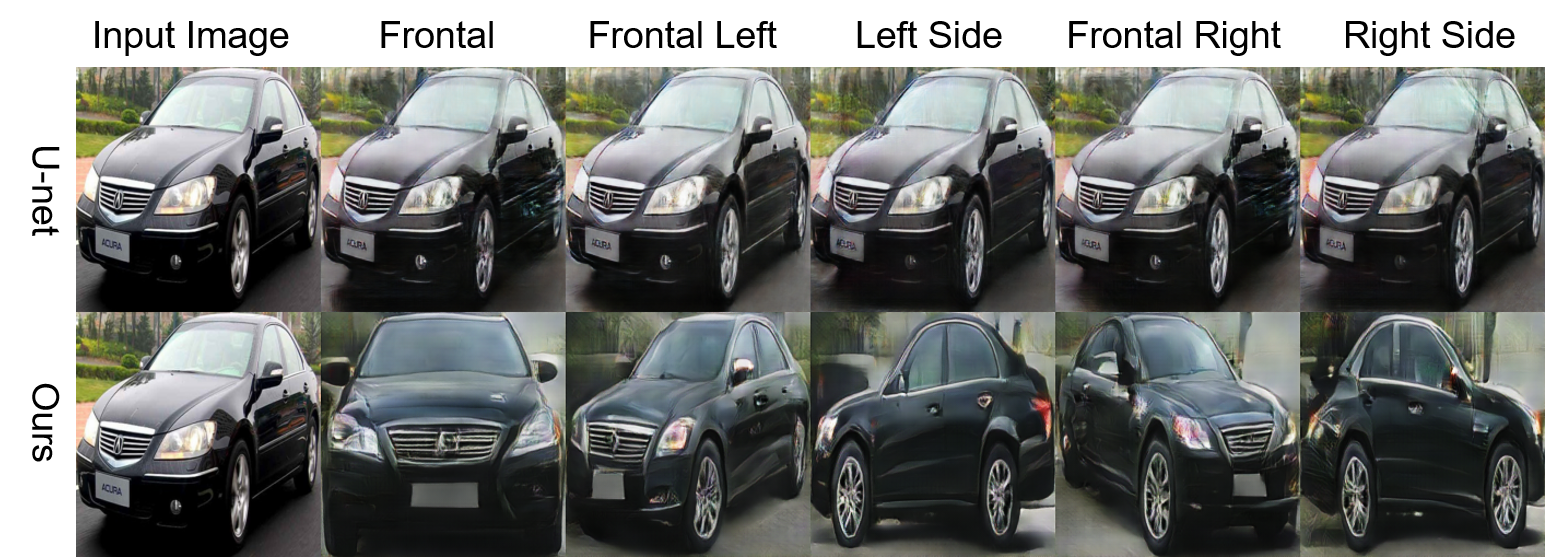}
    \caption{Images generated by U-net (top) and our model (bottom). The first column shows the input image, and the rest columns are images generated with five different viewpoints as condition. Our model generates images with correct viewpoints while U-net fails to accomplish the task.}
    \label{fig:ablation_study}
    %\vspace{-0.15in}
\end{figure}

%\vspace{-0.05in}
\subsection{Ablation Study}

%\textbf{Attention Visualization.} We visualize the non-local attention between the decoder feature and the encoder feature. Specifically, we select one feature point at the decoder feature, then visualize the attention map of the encoder feature map with respect to that point. The attention map has the same spatial size as the encoder feature map. We resize the attention map to the size of the input image for better visualization. As is shown in Fig. [], the point in the decoder feature map covers a certain region in the output image, and the attention map successfully finds out the corresponding regions in the input image. The results demonstrate the effectiveness of our proposed non-local attention connection. 

We further analyze how each part of our model contributes to the overall performance. Specifically, we conduct the identity preservation experiment with the following versions of our model on CompCars dataset: 1) The vanilla model without constrained nonalignment connection (CNC) nor adaptive identity modulation (AIM). The vanilla model shares a similar architecture as DR-GAN. The generator has an encoder-decoder architecture (removing all the AIMs and CNCs), while the discriminator remains the same as our full model. 2) The vanilla model with deformable convolution \cite{dai2017deformable} applied on the $28\times28$ feature block instead of the original convolution. 3) Model with unconstrained nonalignment connection, denoted as ``Global-NC''. Global-NC is a variant of CNC which modifies Eq. \eqref{eq:attention} to search over all the spatial locations in $K$, instead of merely searching a neighborhood region. 4) Model with only CNC. 5) Our model with CNC and Identity Modulation using Eq. (\ref{eq:im}). 6) Our full model with both CNC and AIM using Eq. (\ref{eq:aim}). The discriminator and the loss functions remain unchanged. We also study how the location of CNC influences the final performance. Therefore, we use CNC/Global-NC to connect convolution blocks with different spatial sizes. Specifically, as the structure of the encoder and the decoder in our model is symmetrical to each other, we choose to connect one block in the encoder with the corresponding symmetrical block in the decoder. We apply CNC and Global-NC on feature maps with a $28\times28$ or $56\times56$ spatial resolution. 

Results are shown in Table \ref{table:ablation}. Compared to the vanilla model, using deformable convolution benefits the performance. However, our model with CNC still outperforms deformable convolution. CNC significantly improves the performance of the model compared to Global-NC model and vanilla model by a large margin, demonstrating its effectiveness. Applying CNC to different feature blocks can influence the performance of the model. AIM also makes significant contributions to improving the identity preservation ability of the model. AIM also consistently outperforms IM (Eq. (\ref{eq:im})).

\noindent\textbf{CNC versus Skip-Connection.}  We further analyze how constrained nonalignment connection is crucial to the success of fine-grained image transformation with large geometric deformation. On CompCars dataset, we compare our model with a counterpart, which uses a U-net as the generator with skip-connections to link the encoder and decoder. The other settings of the U-net model remain the same as our model. Fig. \ref{fig:ablation_study} shows the images generated by our model and the U-net model. Unsurprisingly, U-net model ignores the target viewpoint condition and generates images that are almost the same as the input image without changing the views. \textit{Note that duplicating the input image can easily preserve the identity of the input image, but will not provide useful information for the visual recognition systems.} On the contrary, our model can generate identity-preserved images with correct viewpoints, demonstrating the superiority of the our constrained nonalignment connection over skip-connection. 

%For more ablation study on model components and hyper-parameters, please see the supplementary material.

%\vspace{-0.1in}
\section{Conclusion}
We study fine-grained image-to-image transformation with the goal of generating identity-preserved images that can boost the performance of visual recognition and few-shot learning. In particular, we adopt a GAN-based model that learns to encode an image to an output image with different viewpoints as conditions. To better maintain the fine-grained details and preserve the identity, we propose constrained nonalignment connection and adaptive identity modulation, which are demonstrated effective in our extensive experiments on the large-scale fine-grained CompCars and Multi-PIE datasets. Our model outperforms the state-of-the-art image transformation methods in identity preservation and data augmentation for few-shot learning tasks. 

\section{Acknowledgement}
This work is supported in part by NSF awards \#1704337, \#1722847, \#1813709, and our corporate sponsors.

{
\bibliographystyle{ieee}
\bibliography{ms}
}

\end{document}